# JobArabi: An Arabic Corpus and Analysis of Job Announcements from Social Media

**Wajdi Zaghouani**[1], **Shimaa Amer Ibrahim**[1], **Mabrouka Bessghaier**[1], **Houda Bouamor**[2]
[1] Northwestern University in Qatar
[2] Carnegie Mellon University in Qatar
{wajdi.zaghouani, shimaa.ibrahim, mabrouka.bessghaier}@northwestern.edu
hbouamor@qatar.cmu.edu

## Abstract

This paper introduces JobArabi, a large-scale corpus of Arabic job announcements collected from social media between January 2024 and October 2025. The dataset contains 20,528 public posts from X and captures more than two years of employment-related discourse across Arabic-speaking online communities. The corpus was compiled using a linguistically informed query framework covering 21 Arabic keyword families that reflect gendered, plural, formal, and dialectal expressions of recruitment language. The resulting dataset includes posts from institutional, commercial, and individual accounts and provides metadata such as timestamps, engagement indicators, and geolocation when available, enabling temporal and regional analysis of employment discourse. Quantitative analysis reveals several sociolinguistic patterns in online recruitment, including the persistence of gendered hiring language, regional variation in occupational demand, and the emotional framing of recruitment messages. These findings highlight the potential of Arabic social media as a resource for studying labor market communication and linguistic change. The JobArabi corpus, together with documentation and collection scripts, will be released to support research in Arabic NLP, computational social science, and digital labor studies.

## 1. Introduction

The last decade has witnessed profound digital transformation across the Middle East and North Africa (MENA). Social media platforms have matured into critical infrastructure for commerce, information dissemination, and professional life (Kemp and DataReportal, 2025a,b). Within this ecosystem, X (formerly Twitter) serves as a real-time public square where government entities, private corporations, and the general public intersect (Tufekci, 2017; Bruns and Burgess, 2011; Mergel, 2013).

Job recruitment is a window into labor market dynamics. Job announcements reveal labor market demand, in-demand skills, salary benchmarks, and evolving corporate cultures (Squicciarini and Nachtigall, 2021a; Sleeman and team, 2021). For Arabic-speaking labor markets, recruitment discourse offers unprecedented visibility into regional economic dynamics, sectoral growth, and workplace practices.

However, a critical resource gap limits Arabic NLP and CSS research. While English language studies on LinkedIn are abundant and general purpose Arabic corpora exist, large-scale, domain specific corpora for the Arabic job market remain limited (Alyafeai et al., 2022; Altaher et al., 2022; El-Haj, 2025). This gap constrains three research directions: (1) developing NLP models for recruitment discourse; (2) analyzing trends in the Arabic labor market over a two-year period; and (3) investigating sociolinguistic phenomena such as gender bias in recruitment.

We present a corpus of Arabic job announcements from X/Twitter: **JobArabi**, containing **20,528 mentions** from January 2024 to October 2025. Our approach prioritizes three principles: **(1) linguistic diversity** capturing formal, informal, and gender inflected recruitment language; **(2) scale and temporal coverage** enabling robust temporal analysis; and **(3) transparency**ensuring reproducibility.

Our contributions are fourfold:

1. **Novel large-scale corpus:** A two-year dataset of Arabic job posts supporting NLP and CSS tasks.
2. **Robust collection methodology:** A 21 keyword query tailored to Arabic recruitment language, moving beyond simple keyword translation.
3. **Multi-faceted analysis over time:** Temporal, geographic, thematic, engagement, and linguistic perspectives on the Arabic job market.
4. **Data driven emotional framework:** Fine grained emotions revealing sociolinguistic tensions around inclusivity and fairness.

We review related work (Section 2), detail data collection and linguistically aware query design (Section 3), present the core analysis (Section 4), and outline applications and future work (Section 5) before concluding (Section 6).

## 2. Related Work

The JobArabi corpus sits at the intersection of three areas: Arabic corpus linguistics for social media, NLP for job market analysis, and computational social science (CSS) in the MENA region.

### 2.1. Arabic Corpora for Social Media Analysis

Arabic social media resources have grown substantially, with corpora for sentiment (e.g., ASTD) (Nabil et al., 2015), dialect identification (e.g., ArSAS) (Elmadany et al., 2018), abusive/hate speech (Mubarak et al., 2017; Charfi et al., 2024c), irony (Charfi et al., 2024d), and stance detection (Charfi et al., 2024b). These social media corpora have been developed to support a variety of NLP tasks, including linguistic variation and user demographic analysis (Zaghouani and Charfi, 2018), hope and hate emotion detection (Biswas et al., 2025), and hate speech detection (Charfi et al., 2024a).

Domain specific resources, however, remain scarce. While news (e.g., GALE Arabic) and formal text (e.g., OpenSubtitles) are available (Friedman et al., 2012; Lison and Tiedemann, 2016), corpora for *professional* discourse on social media are limited. General purpose datasets often miss recruitment specific terminology (e.g., “Bachelor’s degree,” “entry level,” “financial analyst,” “full time”). JobArabi addresses this gap as a large-scale, social media native corpus focused on the labor market; related off platform work such as *ArabJobs* aggregates job board postings (El-Haj, 2025).

### 2.2. NLP for Job Market Analysis

A substantial English language literature analyzes job postings from LinkedIn, Indeed, and Twitter (Verma et al., 2021; Sran, 2025; Squicciarini and Nachtigall, 2021b; Campion and Campion, 2020), typically covering:

- **Information extraction (IE):** NER driven parsing of titles, organizations, locations, skills (e.g., “Python,” “PMP”), degrees, and experience.
- **Trend analysis:** Tracking demand for roles (e.g., “data scientist”), emergent skills (e.g., “remote work”), and sectoral shifts.
- **Bias and fairness:** Detecting gendered, racial, or ageist language (e.g., “salesman,” “rockstar developer,” “digital native”) (Gaucher et al., 2011).

Our work brings these well studied analyses to Arabic, where fairness and bias in recruitment remain underexplored. The corpus is designed for IE, trend tracking, and bias auditing in Arabic job discourse. Recent surveys and infrastructures (e.g., skill extraction surveys; Nesta’s Open Jobs Observatory) illustrate how postings support near real time labor market intelligence (Senger et al., 2024; Sleeman and team, 2021).

### 2.3. Computational Social Science in the MENA Region

Social media data underpins CSS across MENA, from political sentiment and social movements to information diffusion and public health (Lee et al., 2023; Howard et al., 2011; Lotan et al., 2011). While this work establishes social media as a real time sensor, economic modeling via postings remains less developed. The two-year scope of JobArabi supports analyses of labor dynamics, internal migration (via job locations), and sectoral growth over time.

JobArabi bridges these domains by providing a domain specific Arabic corpus for NLP tasks, grounded in job market analysis, and purpose built to answer CSS questions about labor markets in the Arabic speaking world.

## 3. Methodology and Corpus Construction

### 3.1. Data Collection Framework

We collected data from the X/Twitter platform covering **January 1, 2024, to October 20, 2025** [1]. Collection relied on a keyword tracking query that captured all public original posts and replies containing recruitment related terms. This section details the query design and resulting corpus.

### 3.2. Query Design: A Linguistically Aware Approach

Direct translations of English terms like ”hiring” or ”job” capture only a fraction of how recruitment is actually expressed in Arabic. The language of job announcements varies across formality registers (formal corporate language vs. colloquial postings), employs multiple synonyms ,4..::1و)

[1] The dataset was collected through the Meltwater media monitoring platform: https://www.meltwater.com/en/products/media-monitoring

(J..� 4�}; ("job", "job opportunity"), uses gender inflected forms (4..1y,. vs. 01y,.) ("male employee" vs. "female employee"), and includes time sensitive markers (ًرارy; ,وريy;) ("immediate", "immediately"). A naive English to Arabic keyword translation would yield low recall and miss substantial portions of the market.

**Our solution: A 21 keyword query designed for linguistic diversity.** To ensure high recall while capturing the full range of Arabic recruitment language, we constructed a linguistically aware query accounting for five key dimensions: formality levels, synonymy, common collocations, pluralization, and gender inflection.

The complete Arabic query used is:

> (4..::1و إ\Jcن OR 07I1و إ\Jc;ات OR إ\Jcن 0::1yr OR 0::1yr إ\Jc;ات OR LI ة✓-I� 4..::1و OR LI ة✓-I� 07I1و OR ىꟺꟺ J..� 4�}; OR ىꟺꟺ J..� ص}; OR LI J ll ب_lk.,. OR ب_lk.,. LI 01y,. OR LI 4..1y,. ب_lk.,. OR J ... ll ب_lk.,. ىꟺꟺ OR LI 4ناو,. 07I1و OR LI وري; 0::1yr OR LI وري; (;::::..r OR ىꟺꟺ J..� ض✓ OR 4�}; ة✓-I� J..� OR ة✓-I� J..� ص}; OR ب_lk.,. LI 0::1yr OR LI ًرارy; J ... ll ب_lk.,. OR ب_lk.,. LI ارy; J ... ll)

The Corresponding English Translation:

> ("job announcement" OR "job announcements" OR "recruitment announcement" OR "recruitment announcements" OR "vacant position in" OR "vacant positions in" OR "job opportunity at" OR "job opportunities at" OR "wanted to work in" OR "employee wanted in" OR "female employee wanted in" OR "wanted to work for" OR "available jobs in" OR "immediate hiring in" OR "immediate appointment in" OR "job offer at" OR "vacant job opportunity" OR "vacant job opportunities" OR "hiring required in" OR "urgently hiring in" OR "urgent hiring in")

**Design rationale across five linguistic dimensions:**

- **Formality spectrum:** Spans from formal corporate announcements (0::1yr إ\Jcن "recruitment announcement") to colloquial postings (J ll ب_lk.,. "hiring" / "wanted for work"), capturing recruitment across different organizational contexts and audiences.
- **Synonymy and collocations:** Includes semantically overlapping recruitment phrases (J..� 4�}; "job opportunity", ة✓-I� 4..::1و "vacant position", J..� ض✓ "job offer") that are used interchangeably by employers, preventing missed announcements due to lexical choice.
- **Pluralization:** Covers both singular (إ\Jcن 4..::1و "job announcement") and plural forms (07I1و إ\Jc;ات "job announcements"), capturing recruitment posts referring to either single positions or multiple vacancies.
- **Gender inflection:** Explicitly includes gendered forms (01y,. ب_lk.,. "male employee wanted" vs. 4..1y,. ب_lk.,. "female employee wanted"), enabling subsequent sociolinguistic analysis of gender-specific hiring practices, which represents a central contribution of this work.
- **Immediacy markers:** Includes urgent hiring signals (وري; 0::1yr "immediate hiring", ًرارy; "immediately"), helping distinguish time-sensitive recruitment from standard job postings.

This linguistically aware design ensures the corpus captures the full diversity of Arabic job announcement language and enables analysis of recruitment practices across formal registers and social dimensions.

## 3.3. Corpus Statistics and Scale

The final corpus comprises **20,528 mentions** collected across two periods:

1. **2024 dataset: 8,316 mentions** from January 1-December 31, 2024, comprising both raw post level data and daily aggregated trend metrics.
2. **2025 dataset: 12,212 mentions** from January 1-October 20, 2025, accessed via analytics dashboard providing aggregated topic, location, and keyword snapshots.

**Rapid market expansion evident in corpus growth.** The 2025 data surpassed the entire 2024 volume by **46.8%** within just ten months (8,316 → 12,212 mentions), signaling rapid adoption of X/Twitter for recruitment by employers and job seekers in the Arabic speaking world. This acceleration underscores both the timeliness of this resource and the platform's emerging role as critical labor market infrastructure in the MENA region.

## 3.4. Public Release and Licensing

The dataset will be released upon request through an open-access repository [2]. All materials will be shared under a Creative Commons Attribution (NonCommercial ShareAlike 4.0 license), with preprocessing scripts and model hosted on GitHub for reproducibility.

[2] Interested researchers may apply for access through the following form https://tinyurl.com/4ke5jwyw

## 4. In-Depth Corpus Analysis

This section presents a comprehensive, multi faceted analysis of the JobArabi corpus, using analytics provided by Meltwater pltaform. We examine the Arabic job market through four interconnected lenses: temporal dynamics revealing when recruitment peaks; geographic and sectoral landscape showing where and what jobs dominate; visibility and reach mechanisms that amplify recruitment signals; and sociolinguistic and emotional discourse embedded in recruitment language.

### 4.1. Temporal Trend Analysis (2024-2025): The Rhythm of Recruitment

**Macro level growth and geographic drivers.** The corpus shows **46.8% year over year growth**, with mentions rising from 8,316 in 2024 to 12,212 in the first ten months of 2025. Qualitative insights point to a surge driven by adoption among companies and government entities in the Gulf, particularly **Saudi Arabia and Kuwait**.

**Seasonal hiring cycles reveal shifting corporate calendars.** Temporal patterns differ meaningfully between years:

- **2024:** A quiet first half followed by a third quarter peak (July-September), consistent with post summer budget cycles and seasonal hiring in education and hospitality.
- **2025:** An inverted pattern, with strongest activity in **January-March**, suggesting alignment with formal corporate fiscal calendars and new annual budgets.

**Weekly rhythm indicates predominantly formal, corporate activity.** Mentions peak midweek (Tuesday–Thursday) and drop on Friday–Saturday, corresponding to the regional weekend. Figure 1 shows replies and likes mirror these posting peaks, confirming this is primarily corporate communication rather than informal engagement. This consistency suggests the platform functions as an institutionalized recruitment channel.

### 4.2. Geographic and Sectoral Distribution: Mapping the Market

We examine where and what types of jobs dominate. Geographic concentration and sectoral demand reveal regional economic specialization.

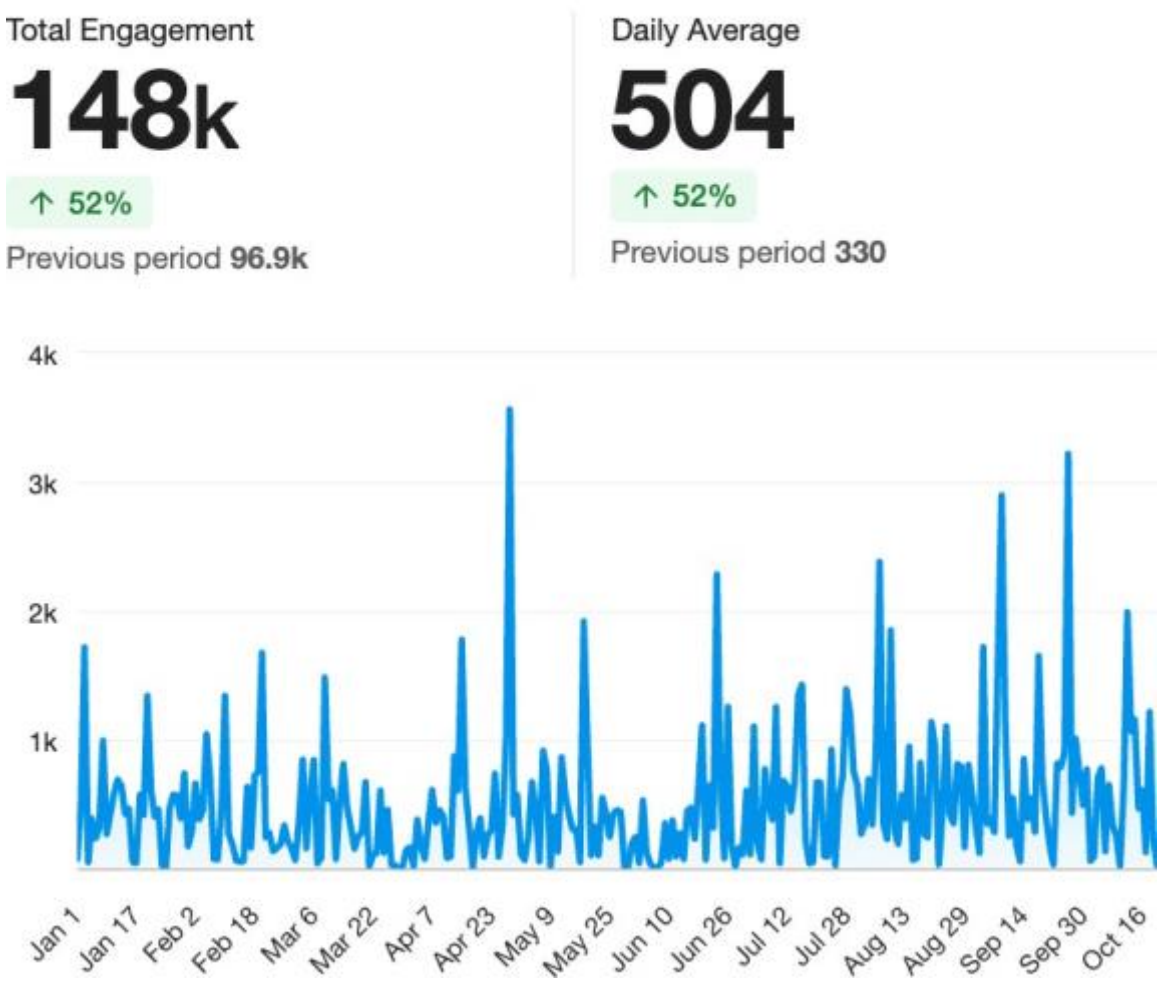

Figure 1: Engagement trends showing midweek peaks in posting volume and user interaction.

#### 4.2.1. Geographic patterns: Concentrated hubs with emerging peripheries

The labor market is highly concentrated geographically, yet shows important variation.

**Saudi Arabia is the undisputed center.** Saudi Arabia accounts for **3,210 mentions or 71.21%** of geo tagged posts. The top five countries (2025) are included Saudi Arabia (3,210), United Arab Emirates (452), Kuwait (262), Egypt (232), Oman (191).

**City level patterns reveal economic specialization.** Within Saudi Arabia, three major hubs emerge, each anchored to different economic sectors:

- **Riyadh (787 mentions):** Administrative capital; recruitment aligns with “Law and Government” (617) and “Finance” (147), reflecting government bureaucracy and financial services.
- **Jeddah (202 mentions):** Commercial hub; postings align with “Business and Industrial” (1,349), including logistics and international trade.
- **Eastern Province/Dammam (178/90 mentions):** Industrial heartland; recruitment ties to energy sector and industrial operations.

**Emerging regional and international trends signal expanding markets.** Smaller cities show distinctive sectoral signatures: Al Ahsa (56), Abha (37), and Taif (33) indicate emerging recruitment in agriculture and tourism. Mentions from the **United**

**States (58)** and **United Kingdom (10)** suggest bidirectional flows between MENA and Western employers.

#### 4.2.2. Sectoral demand: Identifying in demand sectors

Sectoral demand reinforces geographic patterns and reveals where the Arabic job market expands.

**Jobs and Education dominates but is primarily direct listings.** The top five sectors (2025 mentions) are:**Jobs and Education** (4,798),**Business and Industrial** (1,349),**Law and Government** (617),**Health** (457), **People and Society** (441).

Critically, **97.8% of job related posts (4,744 of 4,879)** are direct **"Job Listings"**, with only 2.2% (135) classified as "Career Resources and Planning." This confirms JobArabi captures *active recruitment*, not career advice.

**Sectoral differences reflect regional economic structure and modernization.** Cross referencing sectors with geographic patterns:

- **Public sector ("Law and Government", 617):** Heavily concentrated in Riyadh; includes entities such as the l.:i"1yll :;S"}..1l ("National Center") and various ministries, reflecting government hiring campaigns.
- **Private sector ("Business and Industrial", 1,349):** Distributed across commercial hubs; includes major firms such as Almarai and the Saudi Arabian Railway Company.
- **Role diversity and modernization:** Remote roles, creative positions, and administrative jobs alongside traditional industrial and governmental postings suggest a market increasingly offering non traditional, flexible work arrangements.

### 4.3. Engagement and Reach Dynamics: How Signals Amplify

**Volume and reach are strongly correlated.** For 2024, daily mention volume and daily potential reach show a **strong positive correlation of 0.78**, confirming higher posting volume translates to greater visibility and engagement.

**Engagement is highly asymmetric, concentrating among major employers.** Posts by major news aggregators and official government accounts achieve thousands of interactions, while posts from small firms receive single digit engagement. A single announcement from a major entity can dominate daily reach, meaning visibility is mediated by institutional prominence.

**Long tail discovery extends the reach window.** Reach persists even when new mentions approach zero, indicating job seekers discover older posts via hashtags, keyword searches, and replies weeks or months after posting. This persistent discoverability extends the effective lifetime of recruitment signals.

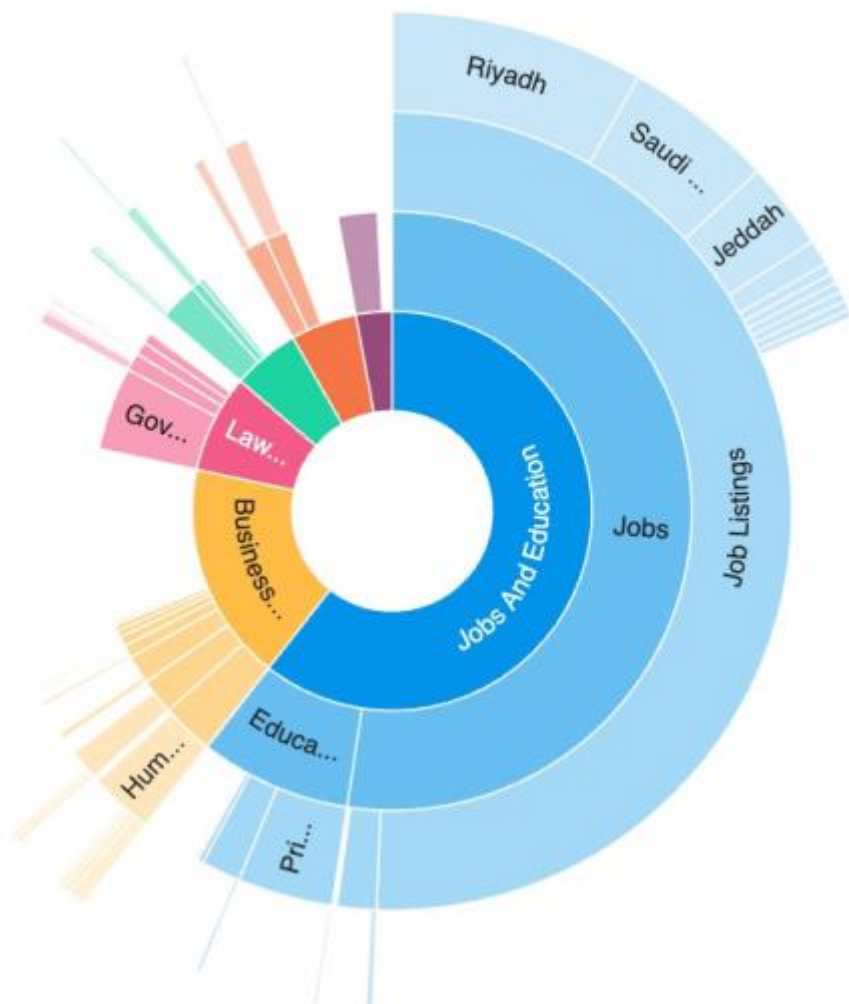

Figure 2: Topic breakdown treemap showing the dominance of direct job listings within the corpus.

### 4.4. Sociolinguistic Analysis: The Language of Recruitment

Keyword, hashtag, and entity data reveal the core lexicon of Arabic recruitment and social norms embedded within it. These linguistic patterns reflect and reinforce labor market practices and values (see Figure 3).

#### 4.4.1. The lexicon of qualification

Professional job requirements follow predictable patterns, establishing baseline expectations.

**Bachelor's degree is the standard entry point.** Keywords س_ر_ll□"lا **"the bachelor's degree" (325 mentions)** and س_ر_ll□□ **"bachelor's degree" (254 mentions)** appear **579 times in total**, indicating that undergraduate education serves as the baseline requirement for many professional roles.

**Mid-level management roles are visible but not dominant.** Demand for supervisory positions is reflected in the keyword **ف..r!..o "supervisor" (87 mentions)**, which appears among the top "Person" entities, indicating active recruitment for supervisory roles.

**Vague language reveals hiring opacity.** Large organizations often rely on catch-all phrases such as **تl.,a.,a□:lا ...;l:;.... "various specializations" (105 mentions)** and **ت□ِl,-1ا ...;l:;....**

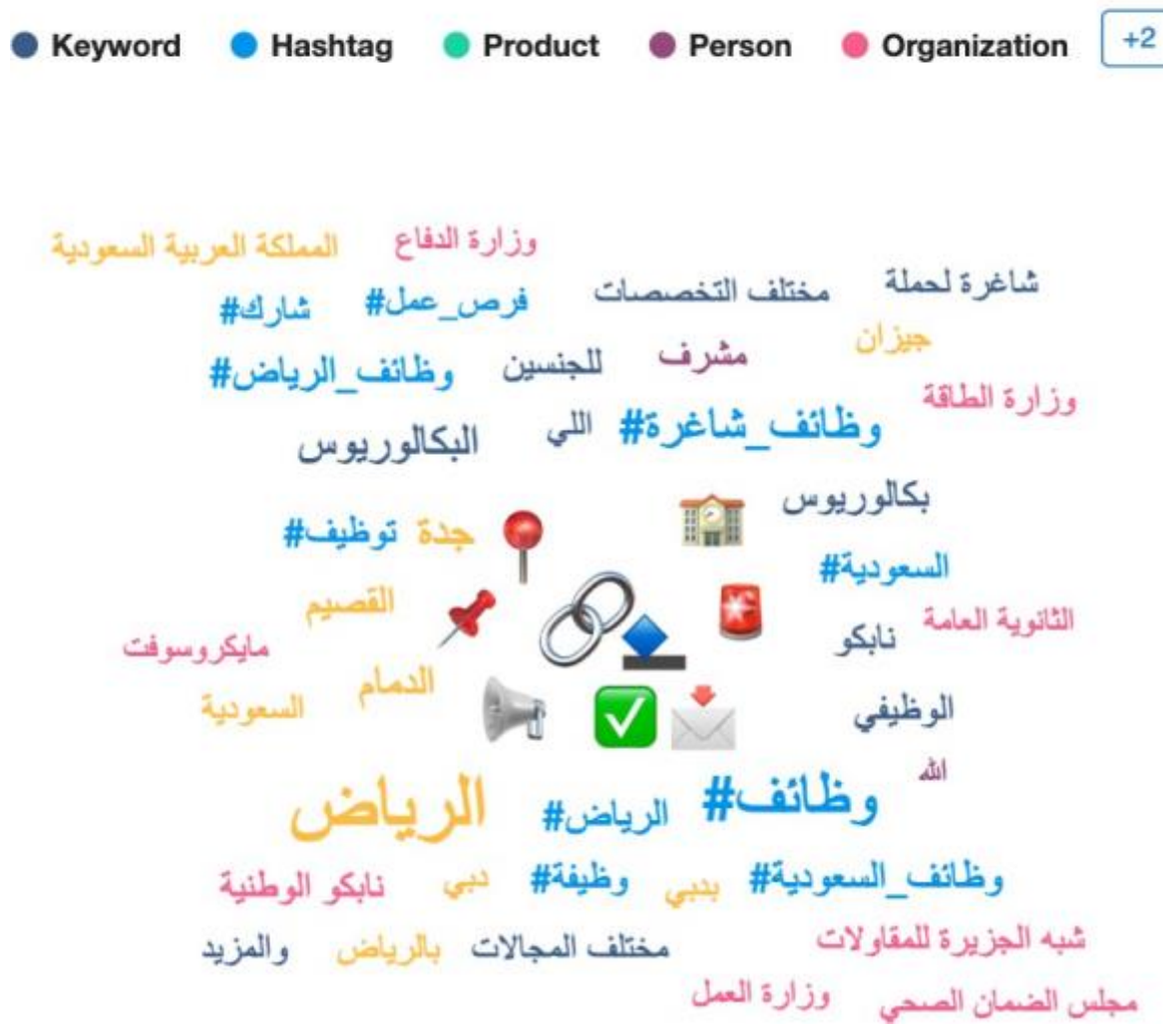

Figure 3: Word cloud visualization of top keywords and entities in the corpus.

**“various fields” (91 mentions)** instead of specifying precise requirements, thereby obscuring the exact qualifications sought.

#### 4.4.2. The hashtag driven ecosystem: How discovery happens

Hashtags function as the primary indexing and discovery mechanism. Job seekers follow hashtags rather than individual employers, organizing their search along topic and location.

**General topic hashtags structure recruitment discourse.** The most frequent hashtags include #07l1و “jobs” (680 mentions), #ة✓-l�_07l1و “vacancies” (343 mentions), and #J..�_ص}; “job opportunities” (115 mentions), reflecting the use of broad tags to reach job-seeking audiences.

**Geographic hashtags enable location-based job search.** Recruitment posts are frequently organized by place using hashtags such as #ض�)ﷺl “Riyadh” (197 mentions), #ض�)ﷺl_07l1و “Riyadh jobs” (131 mentions), #4�ـy...ll “Saudi Arabia” (91 mentions), and #ة..c: “Jeddah” (82 mentions). This geographic granularity facilitates targeted job discovery for location-sensitive candidates.

#### 4.4.3. A corpus in transition: The discourse on gender

**Traditional gendered language persists but is being challenged.** The 2025 keyword data includes **لl□Jll ة_),l, “vacant for men” (62 mentions)** as a notable keyword, highlighting the continued presence of explicitly gendered job postings that restrict access to certain opportunities.

**Inclusive language is more prevalent and associated with positive sentiment.** In contrast, **0,-.i□.,, “for both genders” (79 mentions)** occurs more frequently, indicating a shift toward inclusive recruitment language. Sentiment analysis of keywords also links **0,-.i□.,,** to positive sentiment, suggesting broader community approval of inclusive terminology.

**Community-driven critique reinforces the shift toward inclusivity.** Corpus-level expressions of anger and sadness reflect user-driven critiques of non-inclusive hiring practices, generating social pressure toward more equitable language. Overall, the corpus documents an ongoing sociolinguistic shift in which community sentiment increasingly favors inclusive terminology.

### 4.5. Advanced Emotional and Sentiment Analysis: A Data Driven Framework

Underlying the linguistic patterns lies a rich emotional dimension. Fine grained emotion analysis reveals sociolinguistic tensions and human experiences embedded in recruitment discourse.

The fine grained emotion analysis (Figure 4), combined with 2024 sentiment data and qualitative insights, supports a nuanced, data driven framework for interpreting how the community experiences and responds to the labor market.

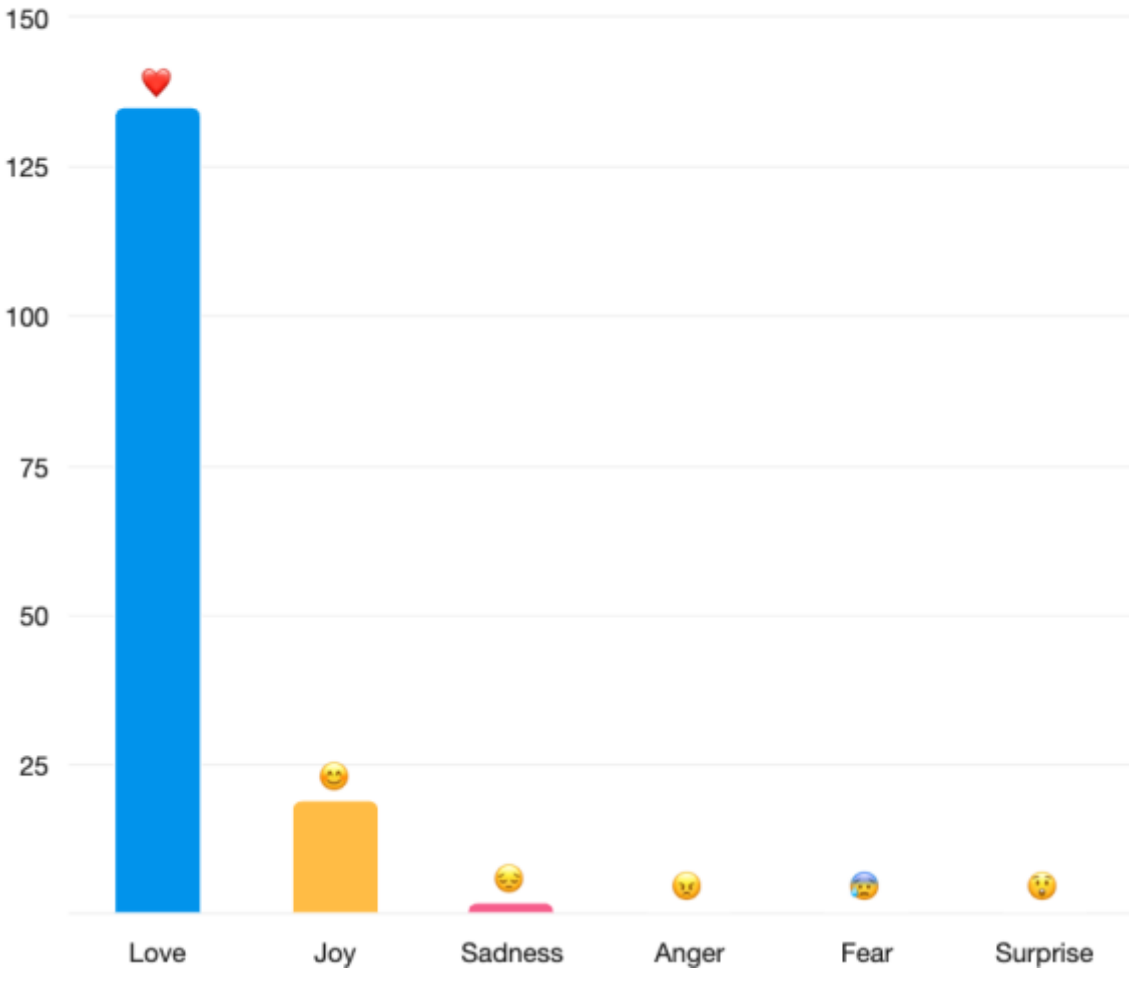

Figure 4: Emotional comparison showing the distribution of Joy, Sadness, and Anger across the corpus.

#### 4.5.1. The baseline: A corpus of formality and neutrality

The 2024 sentiment distribution (Figure 5) establishes the baseline: **89.28% Neutral**. This overwhelming majority reflects the institutional and formal nature of job announcements. Posts are objective and informational ”Company X announces vacancy Y in city Z. Requirements: A, B, C. Apply at link D.” This neutral majority represents the default register of professional recruitment discourse.

**Sentiment**

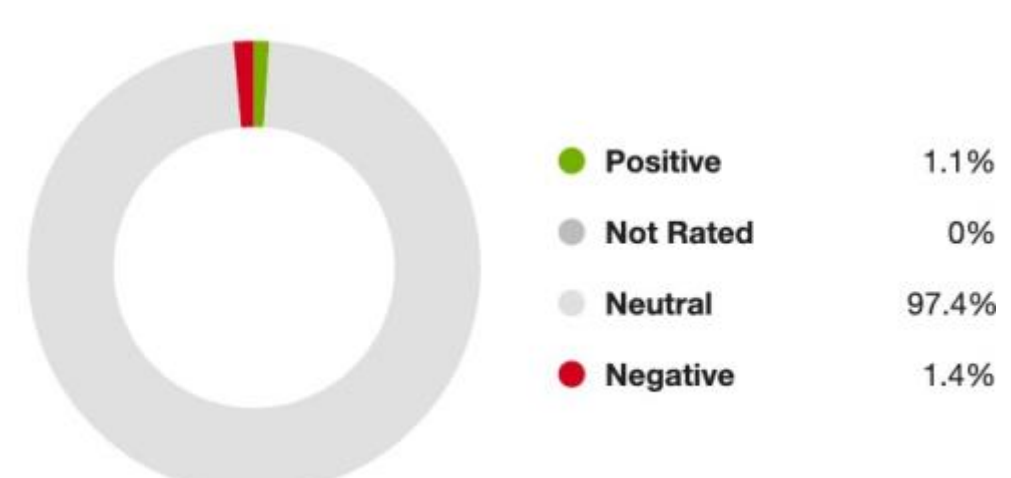

Figure 5: Sentiment distribution showing 89.28% neutral content, reflecting the formal nature of job announcements.

#### 4.5.2. The emotional core: Joy, Sadness, and Anger reveal deeper dynamics

While neutrality dominates, smaller emotional segments reveal key sociolinguistic dynamics.

**The Voice of Aspiration: Joy (3.35% Positive).** Joy aligns with the 3.35% positive portion of the 2024 data, capturing aspirational content and success narratives. High engagement posts often fall here, particularly major announcements from high authority accounts (e.g., ministries announcing large hiring rounds).

**The Voice of Frustration: Sadness and Anger (7.37% Negative).** The 7.37% negative sentiment combines two distinct emotional responses:

- **Anger:** Reflects community pushback against non-inclusive or opaque hiring practices, often directed at explicitly gendered postings such as Jlc:}ll ةُ✓-l� “vacant for men”. In this context, anger functions as a form of linguistic activism, with users publicly criticizing perceived unfair recruitment practices.
- **Sadness:** Captures the lived emotional experience of job seeking-rejections, unemployment stress, and difficulty securing roles. This reflects both individual disappointment and broader structural challenges in the labor market.

**An integrated emotional framework.** Together, these three emotional registers (i.e., **Neutral Formality**, **Joyful Aspiration**, and **Angry/Sad Frustration**) offer an empirically grounded, nuanced interpretation of recruitment discourse.

## 5. Potential Applications and Future Work

The analyses presented in Section 4 reveal rich linguistic and sociolinguistic patterns in Arabic recruitment. JobArabi enables researchers to build on these findings through both immediate applications and longer term research directions.

### 5.1. NLP Model Development

The corpus supports development of recruitment domain NLP models tailored to Arabic.

- **Domain Specific NER:** Train models to extract recruitment specific entities: [JOB_TITLE], [COMPANY], [LOCATION], [SKILL], [CERTIFICATION], [DEGREE], [YEARS_OF_EXPERIENCE], and [SALARY]. This enables structured extraction of job requirements and qualifications from the unstructured recruitment discourse analyzed in Section 4.4.
- **Relation Extraction:** Learn relations among entities; for example, identifying patterns such as [JOB_TITLE: ‘ف_.:.,.‘ “supervisor”] *requires* [DEGREE: ‘وسy� رyllSًِ‘ “bachelor’s de-gree”], thereby uncovering implicit qualification requirements in recruitment discourse.
- **Job Ad Classification:** Build classifiers to categorize jobs by industry, seniority level (entry, mid, senior), and type (full time, part time, remote), supporting fine grained labor market segmentation.

### 5.2. Computational Social Science Applications

The temporal and emotional dimensions analyzed in Section 4 support three key CSS research directions.

- **Real Time Economic Indicators:** Use temporal signals to model labor market velocity, sectoral growth, and economic health as a real time proxy for official reports.
- **Skill Gap Analysis:** Extract required skills from ads and compare with supply side workforce profiles to identify regional skills gaps.
- **Sociolinguistic Monitoring:** Track the evolution of professional Arabic by analyzing changes in gendered terms (e.g., ل‍lc:}ll ö✓-l� "vacant for men") and inclusive expressions (e.g., (;::..i·•1 "for both genders"), and relating these linguistic shifts to patterns of sentiment and emotion in the corpus.

### 5.3. Future Work

Three prioritized directions emerge from current limitations and research opportunities: **Manual Annotation:** Create a gold standard subset of 1,000–2,000 posts with fine grained emotion and bias labels. This supports strong training and evaluation of downstream NLP models, addressing current reliance on automated sentiment and emotion detection. **Algorithmic Bias Detection:** Develop and benchmark models that detect and flag discriminatory or non inclusive language in Arabic job postings. This operationalizes the gender and inclusivity insights from Section 4.4.3, providing actionable tools for employers and job seekers. **Corpus Expansion:** Extend the query to capture informal, dialectal, and emergent recruitment language, including gig work and non formal employment sectors. This addresses current geographic concentration and formality bias in the corpus.

## 6. Conclusion

This paper introduces JobArabi, a corpus of **20,528 Arabic job announcements** from X/Twitter (2024-2025), addressing a critical gap in domain specific Arabic recruitment resources. Our contributions are fourfold: (1) a large-scale dataset spanning two years, enabling robust NLP and CSS research; (2) a **21 keyword linguistically aware query** capturing formal, informal, and gender inflected recruitment language; (3) multi faceted analysis revealing temporal, geographic, and sectoral patterns; and (4) a data driven emotional framework exposing sociolinguistic tensions around inclusivity.

Key findings reveal a rapidly expanding market: mentions surged **46.8%** (2024-2025), **97.8%** are direct job listings, and **71.21%** originate from Saudi Arabia. Crucially, recruitment language is shifting toward inclusive terms, with community sentiment favoring fairness. Fine grained emotions reveal frustration and aspiration beneath the **89.28%** neutral baseline.

JobArabi is a timely resource for NLP, computational social science, and sociolinguistics. It enables recruitment specific NLP tasks, real time economic monitoring, sociolinguistic language change tracking, and bias detection addressing the three research directions outlined in the introduction. Beyond technical contributions, JobArabi documents how digital platforms amplify both opportunity and inequality in the Arabic professional sphere.

## 7. Limitations

While the JobArabi corpus provides a novel and timely resource, its methodology and scope present several limitations:

1. **Keyword-Based Query Limitations:** The corpus relies on a curated set of 21 keywords for data collection. While linguistically informed, this method has inherent constraints:
   - **Recall (False Negatives):** The query may miss job announcements that use alternative phrasing, heavy English code-switching (e.g., "We are hiring"), or implicit recruitment language (e.g., "We are expanding our team") without explicit job-related keywords. Such cases may reduce recall and introduce keyword-based bias in the collected data. Future work could mitigate this limitation through query expansion strategies or weak-supervision methods that automatically identify recruitment discourse beyond predefined keywords.
   - **Precision (Corpus Purity):** The query was designed to capture job announcements, but the methodology also collected "replies". This results in the inclusion of a user-driven "meta discourse" of frustration, anger, and sadness. While this became a key sociolinguistic finding, it means the corpus contains a mix of formal job postings and user reactions rather than exclusively the former.
2. **Platform Bias (X/Twitter):** The corpus is sourced exclusively from the X/Twitter platform. The findings therefore reflect the labor market dynamics, user demographics, and linguistic norms specific to this environment. As a result, the dataset primarily captures recruitment practices occurring in social-media contexts and may not fully represent the broader Arabic job market, which also operates through dedicated professional platforms (e.g., LinkedIn, Bayt.com) and corporate career portals.

3. **Geographic Concentration (GCC Focus):** The corpus is geographically concentrated in the Gulf Cooperation Council (GCC) region. In particular, 71.21% of geo-tagged posts originate from Saudi Arabia, reflecting the strong presence of social media recruitment in that market. While the dataset includes posts from other Arabic-speaking regions, these appear in smaller proportions. Consequently, the corpus should be interpreted primarily as a Saudi/Gulf-centered view of Arabic recruitment discourse rather than a comprehensive representation of the broader MENA labor market.

4. **Text Centric Analysis:** The paper's analysis focuses on textual data, including keywords, sentiment, and emotion. Job announcements on a visual platform like X/Twitter often rely on images or videos containing embedded text. The methodology does not specify if these multimodal elements were processed (e.g., via OCR), potentially missing data from image based job ads.

5. **Temporal Data Heterogeneity (2024 vs. 2025):** The dataset combines two forms of access to the Meltwater platform. The 2024 component is based on raw social media posts retrieved through keyword queries, while the 2025 component relies on aggregated analytics and trend metrics generated by the Meltwater dashboard. These two access modes differ in terms of available fields and levels of aggregation. As a result, certain measurements may be influenced by platform-specific analytics methods, which could introduce minor measurement inconsistencies between years.

6. **Multimodality Gap:** Many recruitment posts on social media appear as image-based advertisements or graphical posters containing embedded text. The current corpus focuses on textual content extracted from posts and does not apply Optical Character Recognition (OCR) to process text contained in images.

## 8. Ethics Statement

The collection and analysis of public data, especially concerning sensitive topics like employment and discrimination, require careful ethical consideration.

1. **Data Sourcing and Privacy:** All data was collected from the *public* X/Twitter platform, including both original posts and replies. No private data was accessed. However, job seeking is a sensitive activity. The corpus contains expressions of user frustration, including "Sadness" (potentially related to unemployment) and "Anger" (related to perceived discrimination). The aggregation of this public data into a research corpus creates a potential re identification risk, even if user handles are pseudonymized.

2. **Potential for Misuse (Dual Use):** The resource, particularly the analysis of user frustration, could be misused. Malicious actors could potentially use this data to target frustrated or vulnerable job seekers with misinformation or scams. Unscrupulous recruiters could analyze the "Voice of Frustration" to identify and exploit desperate candidates.

3. **Content and Researcher Harm:** The corpus contains content that is inherently sensitive. This includes user driven "meta discourse" criticizing non inclusive or opaque hiring practices and traditional, gender specific job ads. Researchers analyzing this data may face psychological harm from repeated exposure to content reflecting discrimination, anger, and sadness.

4. **Mitigation and Justification:** We believe the benefit of this corpus for enabling research into algorithmic fairness, bias, and sociolinguistic shifts in the Arabic job market outweighs the risks. To mitigate these risks, the dataset will be released only to vetted researchers for non commercial purposes under a strict data use agreement. All personally identifiable information (PII), such as user handles and exact user IDs, will be anonymized or removed from the public release. We urge all researchers using this data to handle it with care and respect for the individuals represented within it.

5. **Platform bias:** Because the dataset is collected from X/Twitter, it reflects the demographics, communication styles, and platform dynamics of this environment, which may not represent the broader Arabic labor market.

6. **Emotion analysis limitation:** Emotion-analysis results should also be interpreted with caution, as automatic models may misinterpret context, sarcasm, or dialectal variation in Arabic.

## Acknowledgments

This work was made possible by the National Priorities Research Program (NPRP) grant NPRP14C-

0916-210015 from the Qatar National Research Fund (QNRF), a member of the Qatar Research, Development and Innovation Council (QRDI).